\documentclass{article}
\usepackage{hyperref}
\usepackage{url}

\usepackage{amssymb,amsmath,amsthm}

\usepackage{algorithmic} 
\usepackage{algorithm} 
\usepackage[T1]{fontenc}
\usepackage{graphicx}
\usepackage{graphics}
\usepackage{multirow}
\usepackage{fancyhdr}
\usepackage{bm}
\usepackage{subfigure}
\usepackage{authblk}

\usepackage{pgfplots}
\usepackage{etex}
\reserveinserts{18}
\usepackage{morefloats}
\usepackage{array}
\usepackage{colortbl}

\usepackage{atveryend}
\makeatletter
\let\origcitation\citation
\AtEndDocument{\def\mycites{}%
  \def\citation#1{\g@addto@macro\mycites{#1^^J}\origcitation{#1}}}
\AtVeryEndDocument{\newwrite\citeout\immediate\openout\citeout=\jobname.cit
  \immediate\write\citeout{\mycites}\immediate\closeout\citeout}
\makeatother

\usepackage{tikz}
\usepackage{tikz-3dplot}
\usetikzlibrary{fadings,calc}

\tikzfading[name=radialfade, inner color=transparent!0, outer color=transparent!100]

\DeclareMathOperator*{\argmax}{argmax}
\DeclareMathOperator*{\argmin}{argmin}

\newcommand{\cO}{\ensuremath{\mathcal{O}}}
\newcommand{\cI}{\ensuremath{\mathcal{I}}}

\newcommand{\R}{\ensuremath{\mathbb{R}}}

\newcommand{\e}{\mbox{e}}
\newcommand\T{\rule{0pt}{2.3ex}}
\newcommand\B{\rule[-1.0ex]{0pt}{0pt}}
\newcommand{\quotes}[1]{``#1''}

\newcommand{\FW}{\mbox{\tiny FW}}
\newcommand{\A}{\mbox{\tiny A}}
\newcommand{\SW}{\mbox{\tiny SW}}

\newtheorem{prop}{Proposition}

\theoremstyle{definition}

\title{A PARTAN-Accelerated Frank-Wolfe Algorithm for Large-Scale SVM Classification}

\author[1]{Emanuele Frandi}
\author[2]{Ricardo \~Nanculef}
\author[3]{Johan A. K. Suykens}
\affil[1,3]{\small ESAT-STADIUS, KU Leuven, Belgium \texttt{\{efrandi,johan.suykens\}@esat.kuleuven.be}}
\affil[2]{\small Department of Informatics, Federico Santa Mar\'ia University, Chile \texttt{jnancu@inf.utfsm.cl}}
\date{}

\begin{document}

\maketitle

\begin{abstract}
Frank-Wolfe algorithms have recently regained the attention of the Machine Learning community. Their solid theoretical properties and sparsity guarantees make them a suitable choice for a wide range of problems in this field. In addition, several variants of the basic procedure exist that improve its theoretical properties and practical performance. In this paper, we investigate the application of some of these techniques to Machine Learning, focusing in particular on a Parallel Tangent (PARTAN) variant of the FW algorithm that has not been previously suggested or studied for this type of problems. We provide experiments both in a standard setting and using a stochastic speed-up technique, showing that the considered algorithms obtain promising results on several medium and large-scale benchmark datasets for SVM classification.
\end{abstract}

\section{Introduction}\label{intro-sec}

The Frank-Wolfe algorithm (hereafter FW) is a classical method for convex optimization that has seen a substantial revival in interest from researchers \cite{wolfe1954,Jaggi2013ICMLa,harchaoui14}. Recent results have shown that the family of FW algorithms enjoys powerful theoretical properties such as iteration complexity bounds that are independent of the problem size, provable primal-dual convergence rates, and sparsity guarantees that hold during the whole execution of the algorithm  \cite{clarkson_coresets,Jaggi2013ICMLa}. Furthermore, several variants of the basic procedure exist which can improve the convergence rate and practical performance of the basic FW iteration \cite{wolfe1970,GuelatMarcotte,SWAP_paper,jaggi_linconv}. 
Finally, the fact that FW methods work with projection-free iterations is an essential advantage in applications such as matrix recovery, where a projection step (as needed, e.g., by proximal methods) has a super-linear complexity \cite{Jaggi2013ICMLa,SSCI}. As a result, FW is now considered a suitable choice for large-scale optimization problems arising in several contexts such as Machine Learning, statistics, bioinformatics and other fields \cite{signoretto14book,IJPRAI11,Jaggi2013ICMLb}. 
In the context of SVM classification, for example, FW methods have been shown to perform well on large-scale datasets with hundreds of thousands of examples, thus providing a promising alternative to solvers such as Active Set methods and SMO \cite{joachims99,platt99}, whose applicability is often limited to small and medium scale problems \cite{IJPRAI11,SWAP_paper}. 


In this paper, we consider the application of some well-known variants of the FW algorithm to Machine Learning problems, focusing in particular on a type of FW iteration known in the literature as PARTAN, which to the best of our knowledge has not previously been employed for this kind of application. Using several benchmark SVM datasets, we show that this variant is able to accelerate the standard FW method, obtaining on average a $2.52\times$ speedup in CPU time. Furthermore, we show how some FW variants indeed display a faster convergence rate in practice using a primal-dual stopping criterion, though their advantage is limited when the value of the tolerance parameter is not too strict. Finally, to further improve running times on large problems, we consider a random sampling speedup technique, and elaborate on its advantages and drawbacks, particularly on the existence of a tradeoff between iteration complexity and risk of a premature convergence.

\subsection*{Structure of the Paper}

The paper is organized as follows. Section \ref{FW_sec} provides a general overview of the FW method and its modifications, their theoretical properties and some applications to Machine Learning, while in Section \ref{PARTAN_sec} we examine in more detail the PARTAN variant of FW. Then, in Section \ref{exp_sec}, we perform numerical experiments on SVM problems to assess the performance of the considered methods, and close the paper by summarizing our conclusions in Section \ref{conclusions_sec}.

\section{The Frank-Wolfe Method and its Variants}\label{FW_sec}

The FW algorithm \cite{wolfe1954} is a general method to solve optimization problems of the form
\begin{equation}\label{basic_problem}
\min_{\alpha \in \Sigma} \; f(\alpha),
\end{equation}
where $f: \R^m \rightarrow \R$ is a convex differentiable function with Lipschitz continuous gradient, and $\Sigma \subset \R^m$ a compact convex set. The main idea behind the FW iteration is to exploit a linear model of the objective function at the current iterate to define a new search direction. In its basic form, the standard {FW} algorithm can be schematized as in Algorithm \ref{alg:FW}. 

\begin{small}
\begin{algorithm}[!ht]
\begin{algorithmic}[1]
\STATE \textbf{Input:} an initial guess $\alpha^{0}$. %
\FOR{$k = 0,1,\ldots$}
	\STATE Define a search direction ${d}^{(k)}_{\FW} = {u}^{(k)} - {\alpha}^{(k)}$, where 
        \begin{equation}\label{linear_subpb}
        u^{(k)} 
        \in \argmin_{u \,\in\, \Sigma} \, (u-\alpha^{(k)})^T \nabla f({\alpha}^{(k)}). 
        \end{equation}
	\STATE Choose a stepsize $\lambda^{(k)}$, either via the line-search 
	\[
	\lambda^{(k)} \in  \argmin_{\lambda \,\in\, [0,1]}  f({\alpha}^{(k)} + \lambda {d}^{(k)}_{\FW})\,
	\] 
	or with the rule $\lambda^{(k)} = 2/(k+2)$ \cite{Jaggi2013ICMLa}. 
        \STATE Update: 
        \[
        {\alpha}^{(k+1)}=  {\alpha}^{(k)} + \lambda^{(k)} {d}^{(k)}_{\FW} = (1-\lambda^{(k)}){\alpha}^{(k)} + \lambda^{(k)} {u}^{(k)} .
        \] 
\ENDFOR
\end{algorithmic}
\caption{\label{alg:FW} The general FW algorithm.}
\end{algorithm}
\end{small}

\subsection{Theoretical Properties} 

We summarize here, for the sake of completeness, some well-known primal-dual convergence results for the FW algorithm. Proofs for these results can be found in \cite{Jaggi2013ICMLa,lan14}.

\begin{prop}[Sublinear convergence]\label{subconv}
Let $\alpha^{*}$ be an optimal solution for problem (\ref{basic_problem}). Then, for $k \geq 1$, the iterates of Algorithm \ref{alg:FW} satisfy 
\begin{equation*}
f(\alpha^{(k)}) - f(\alpha^{*}) \leq \frac{4C_f}{k+2}\,,
\end{equation*}
where $C_f$ is the \textit{curvature constant} of the objective \cite{Jaggi2013ICMLa}.
\end{prop}


\textbf{Choice of the stopping criterion.} As a consequence of Proposition \ref{subconv}, we immediately have that Algorithm \ref{alg:FW} requires $\cO(1/\varepsilon)$ iterations to obtain an \textit{$\varepsilon$-approximate solution}, i.e. a solution $\alpha^{(k)}$ s.t. $f(\alpha^{(k)}) - f(\alpha^{*}) \leq \varepsilon$. However, given that the \textit{primal gap} $f(\alpha^{(k)})-f(\alpha^{*})$ is not a  computable quantity, this fact cannot be exploited directly. Instead, the stopping condition for FW algorithms is usually based on the following \textit{duality gap} criterion \cite{Jaggi2013ICMLa}:
\begin{equation}\label{duality_gap}
\Delta_d^{(k)} := \max_{u \,\in \,\Sigma} \,(\alpha^{(k)}-u)^T \nabla f(\alpha^{(k)}) \leq \varepsilon \ .
\end{equation}
This is motivated by the fact that the duality gap provides an upper bound for the primal gap, i.e. $f(\alpha^{(k)}) - f(\alpha^{*}) \leq \Delta_d^{(k)}$,  while at the same time enjoying the same asymptotic guarantees.
\begin{prop}[Primal-dual convergence]\label{dualconv}
After $K \geq 2$ iterations, Algorithm \ref{alg:FW} produces at least one iterate $\alpha^{(\bar k)}$, $1\leq \bar k \leq K$, s.t. 
\begin{equation}\label{dual_bound}
\Delta_d^{(\bar k)} \leq \frac{27 C_f}{2(K+2)}\,. 
\end{equation}
\end{prop}
From Proposition \ref{dualconv}, it immediately follows that the $\cO(1/\varepsilon)$ complexity bound holds for $\Delta_d$ as well. Furthermore, the above results give the tolerance parameter $\varepsilon$ a clean interpretation as a tradeoff between optimization accuracy and overall computational complexity.


\subsection{Variants on the Classical Iteration}

Though endowed with solid theoretical properties, the standard FW algorithm exhibits a rather slow convergence rate, and is known to be prone to stagnation, as $d^{(k)}_{\FW}$ tends to become nearly orthogonal to the gradient when nearing a solution \cite{IJPRAI11}. Solutions to this drawback date back to the 1970s, and mostly consist in algorithmic variations where an alternative search direction is added to avoid stalling.

Among the most well-known variants of this kind is the Modified Frank-Wolfe method (MFW) \cite{wolfe1970,GuelatMarcotte,SWAP_paper,jaggi_linconv}. In the modified FW iteration, we define an alternative search direction by \textit{maximizing} the linear model: 
\[
v^{(k)}  \in \argmax_{v \,\in\, \Sigma} \, (v-\alpha^{(k)})^T \nabla f({\alpha}^{(k)}), 
\]
and then setting $d^{(k)}_{\A} =  \alpha^{(k)} - v^{(k)}$. The best descent direction is then selected, i.e. we choose $d^{(k)}_{\A}$ if $\nabla f({\alpha}^{(k)})^T d^{(k)}_{\A} \leq \nabla f({\alpha}^{(k)})^T d^{(k)}_{\FW}$, and stick to the standard $d^{(k)}_{\FW}$ otherwise.

Another option is to use a pairwise (or \quotes{swap}) FW iteration as proposed in \cite{SWAP_paper}, where the alternative search direction is defined as $d^{(k)}_{\SW} =  u^{(k)} - v^{(k)}$. In this case, the choice between $d^{(k)}_{\FW}$ and $d^{(k)}_{\SW}$ is based on a greedy criterion, i.e. we select the step that yields the best function value. It can be proved that the resulting procedure enjoys properties analogous to those of the MFW algorithm.

As the specialization of these algorithms has already been presented extensively in \cite{SWAP_paper}, we do not discuss them further, and refer to the literature for implementation details. In a similar vein, other options for improving the FW iterations, such as conjugate direction based FW or FW with optimization on a 2-dimensional convex hull \cite{stiff-moving}, are not included in this paper due to space constraints.

From a theoretical point of view, these variants often enjoy improved convergence guarantees. In particular, under suitable hypotheses \footnote{We refer to the specialized literature for the detailed analyses \cite{GuelatMarcotte,SWAP_paper,jaggi_linconv}, noting that the necessary hypotheses are satisfied for the test problems used in this paper.}, a linear convergence rate in primal gap can be obtained, i.e. for sufficiently large $k$ we have
\begin{equation*}
\frac{f(\alpha^{(k+1)}) - f(\alpha^{*})}{f(\alpha^{(k)}) - f(\alpha^{*})} \leq M,
\end{equation*}
with $M \in (0,1)$ a constant.


However, to the best of our knowledge, no analogous results improving Proposition \ref{dualconv} were obtained for the duality gap, meaning that there is no \textit{a priori} guarantee that a stopping criterion based on $\Delta_d$ is able to capture the improved behaviour of the algorithm. Furthermore, as the linear convergence results are asymptotic in nature, it is not possible in general to predict whether for a given tolerance the linear rate will kick in before the algorithm stops. Some of the experiments in Section \ref{exp_sec} aim precisely at investigating these issues and their practical impact.

\subsection{Applications to Machine Learning}\label{applications_sec}

One of the most prominent examples of applications of FW-based algorithms to the field of Machine Learning is given by the binary nonlinear $L_2$-SVM training problem \cite{coreSVMs05tsang}:
\begin{equation}\label{svm_problem}
\min_{\alpha \,\in\, \R^{m}} \;\; f(\alpha) = \tfrac{1}{2}  \alpha^{T} K \alpha\ \;\;\;\; \mbox{s.t. } \sum_{i=1}^m \alpha_i = 1,
\, \; \alpha \geq 0 \; ,
\end{equation}
Here, $K$ is a positive definite kernel matrix, and the feasible set $\Sigma$ is the unit simplex, whose vertices are the coordinate vectors $e_{1},\ldots,e_{m}$. It is easy to see that in this case
\begin{equation}\label{svm_vertex}
u^{(k)} = e_{i_{*}^{(k)}}, \qquad \mbox{where } \; i_{*}^{(k)} \in \argmin_{i=1,\ldots,m} \nabla f(\alpha^{(k)})_i .
\end{equation}
Though FW methods can in principle be applied to any SVM formulation giving rise to a compact and convex feasible set, the $L_2$-SVM is chosen here because of its convenience. The geometry of the unit simplex yields indeed very simple formulas for the key steps in the FW iteration, which from a computational perspective leads to an extremely efficient implementation \cite{SWAP_paper}. 

It we denote $\mathcal{I}^{(k)} = \{i\, | \, \alpha^{(k)}_{[i]} > 0 \}$, it follows directly from (\ref{svm_vertex}) that at iteration $k$ the solution can be expressed in terms of at most $k + |\cI^{(0)}|$ data points 
\cite{Jaggi2013ICMLa,SWAP_paper}, or in other words that the number of  Support Vectors is bounded during the entire run of the algorithm, which constitutes a substantial advantage of FW methods in comparison to methods with dense iterates. This holds true in particular for nonlinear SVM problems with datasets where the solution is sparse (in terms of the number of SVs defining the classification model), on which the latter suffer from the so-called \quotes{curse of kernelization} and are unable to recover the sparsity of the solution \cite{curse12}. In addition, Proposition \ref{dualconv} implies that the total number of iterations is independent of the dataset size $m$. Together with the sparsity certificate, this also implies that the memory requirement for the whole algorithm is bounded independently of $m$.

Another related problem that can be tackled with a FW method is the Lasso problem with a $1$-norm constraint
\begin{equation*}
\min_{\alpha \,\in\, \R^{m}} \;\; f(\alpha):=\| A\alpha - b \|_{2}^{2} \;\;\;\; \mbox{s.t. }  \| \alpha \|_{1} \leq t \,,
\end{equation*}
where $A \in \R^{n\times m}$ is a measurement matrix and $b \in \R^{n}$. In this case, the advantage of a FW-based method would be the possibility to well approximate the solution of high-dimensional problems using a reduced set of explanatory variables. Indeed, from Proposition \ref{subconv} we have that at most $\cO(1/\varepsilon)$ \quotes{active} features are required to reach an $\varepsilon$-approximate optimality, independently of the dimensionality of the feature space in which the observations have been embedded.

Finally, matrix recovery problems with nuclear norm regularization of the form
\begin{equation*}
\min_{\alpha \,\in\, \R^{n \times m}} \;\; f(\alpha):=\| \mathcal{A}(\alpha) - b \|_{2}^{2} \;\;\;\; \mbox{s.t. }  \| \alpha \|_{*} \leq t \,,
\end{equation*}
where $\mathcal{A}:\R^{n\times m} \rightarrow \R^{p}$ is a linear operator and $b \in \R^{p}$, have also been successfully tackled with FW-based solvers \cite{SSCI}. The motivation here is mainly that FW methods do not require projection steps. The solution of the linear approximation step can be obtained in a fast way by solving a largest eigenvalue problem, as opposed to proximal methods that require a full SVD of the gradient matrix at each iteration, which is prohibitive for large-scale problems.

As a motivating example, we consider the SVM problem (\ref{svm_problem}) for the experiments in this paper, not only because of its significance, but also to allow for a comparison with the results obtained in previous research efforts \cite{CIARP,Quaderni,IJPRAI11,SWAP_paper,NIPS14}.

\section{PARTAN Frank-Wolfe Iterations}\label{PARTAN_sec}

Another variant of the FW algorithm that has been proposed and successfully employed (for example, in traffic assignment applications \cite{florian1987efficient,LeBlanc,Arezki,stiff-moving}) consists in an adaptation of the method of Parallel Tangents ({PARTAN}) to FW iterations \cite{Luenberger}. To the best of our knowledge, though, this scheme has not yet been investigated in Machine Learning applications. 

\begin{figure}[!ht]
\begin{center}
\begin{tikzpicture}[tdplot_main_coords,scale=0.5,>= stealth,point3/.style = {draw, circle,  fill = red, inner sep = 1.4pt}]

\node[] (a) at (6,0,0) {};
\node[] (b) at (0,6,0) {};
\node[] (c) at (0,0,6) {};
\node[] (o) at (0,0,0) {};

 \filldraw[
        draw=red!0,%
        fill=gray!50,%
    ]          (6,0,0)
            -- (0,6,0)
            -- (0,0,6)
            -- cycle;

\draw [] (6,0,0)--(0,6,0);
\draw [] (0,6,0)--(0,0,6);
\draw [] (0,0,6)--(6,0,0);


    \draw[dashed,->] (0,0,0) -- (7,0,0) node[anchor=south]{};
    \draw[dashed,->] (0,0,0) -- (0,7,0) node[anchor=north west]{};
    \draw[dashed,->] (0,0,0) -- (0,0,7) node[anchor=south]{};

\node[label= {[label distance=-0.0cm]-90:$\!\!{\alpha^{(k-1)}}$} ] (mfw) at ($0.75*(4.26,1.08,3.44)$) [circle,fill=black,scale=0.2] {};
\node[label= {[label distance=0.05cm]90:$\!\!{\alpha}^{(k+1)}$}] (swap) at ($0.75*(0.8,2.4,4.8)$) [circle,fill=black,scale=0.2] {};
\node[label={[label distance=-0.15cm]95:${\tilde \alpha}$}] (fw) at ($0.75*(2.24,1.68,4.08)$) [circle,fill=black,scale=0.2] {};

\node[label= {[label distance=+0.00cm]0:{${\alpha}^{(k)}$}}] (current) at ($0.75*(3.2,2.4,2.4)$) [circle,fill=black,scale=0.4] {};
\draw [thick,dashed, teal] (current)--(c) node [pos=0.8, left, color=black] {} ;
\draw [thick, dashed, teal] (mfw)--(fw) node [right,pos=1.0, color=black] {};

\node[point3,label={[label distance=0.05cm]0:{${u}^{(k)}$}}] at (c) {};

\draw [thick,->] (fw)--(swap);
\draw [thick,->] (current)--(fw);

\end{tikzpicture} 
\end{center}  
\vskip -0.3cm
\caption{\label{partanfig}Sketch of the search directions used by PARTAN iterations.}
\end{figure}
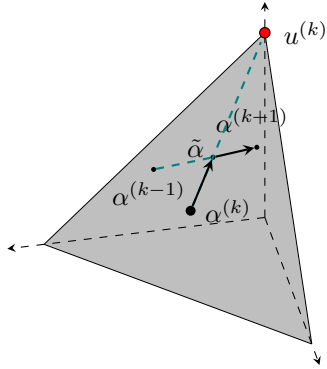

The basic idea, as seen from Figure \ref{partanfig}, is to incorporate previous information by performing an averaging between the classical FW step and the previous iterate. First, an intermediate FW step is defined:
\begin{equation}\label{intermediate_step}
\tilde \alpha = (1-\lambda^{(k)}) \alpha^{(k)} + \lambda^{(k)} u^{(k)}.
\end{equation}
Then, the previous iterate is used to define an extra search direction:
\begin{equation}\label{averaging_step}
\alpha^{(k+1)} = \tilde \alpha + \mu^{(k)} (\tilde \alpha - \alpha^{(k-1)}).
\end{equation}
Stepsizes $\lambda^{(k)}$ and $\mu^{(k)}$ can be determined via line-search.

\tdplotsetmaincoords{55}{168}

\begin{figure}[h!]
\begin{center}
\begin{minipage}[b]{0.425\linewidth}
\begin{tikzpicture}
 [
    scale=0.45,
    >=stealth,
    point/.style = {draw, circle,  fill = black, inner sep = .5pt},
    point3/.style = {draw, circle,  fill = red, inner sep = .8pt},
    dot/.style   = {draw, circle,  fill = black, inner sep = .2pt},
  ]
  
  \node (n1) at (0,0) [point, label = below:$y_{1}\text{$=$}y_{3}$] {};
  \node (n2) at (9,0) [point, label = below:$y_{2}$] {};
  \node (n3) at (4,5) [point, label = {above:$$}] {};

  \draw[-] (n3) -- node (a) [ ] {} (n1);
  \draw[-] (n3) -- node (b) [ ] {} (n2);
  \draw[-] (n2) -- (n1);

   \node (a1) at (4,0) [point3, label = below:$x_{\ast}$] {};
 
   \coordinate (start) at (5,3);
   \node (a1) at (start) [point, label = above:$\tiny{x_{1}}$] {};
   \node (a2) at ($(a1)!0.5!(n1)$) [point, label = above:$x_{2}$] {};
   \node (a3) at ($(a2)!0.45!(n2)$) [point, label = above:$x_{3}$] {};
   \node (a4) at ($(a3)!0.4!(n1)$) [point] {};
   \node (a5) at ($(a4)!0.35!(n2)$) [point] {};
   \node (a6) at ($(a5)!0.3!(n1)$) [point] {};
   
  \draw[-] (a1) -- (a2);
  \draw[-] (a2) -- (a3);
  \draw[-] (a3) -- (a4);
  \draw[-] (a4) -- (a5);
  \draw[-] (a5) -- (a6);
 
  \draw[dotted] (a2) -- (n1);
  \draw[dotted] (a3) -- (n2);
  \draw[dotted] (a4) -- (n1);
  \draw[dotted] (a5) -- (n2);

\end{tikzpicture}
\end{minipage}
\begin{minipage}[b]{0.45\linewidth}
\begin{tikzpicture}
  [
    scale=0.25,
    >=stealth,
    point/.style = {draw, circle,  fill = black, inner sep = .5pt},
    point2/.style = {draw, circle,  fill = black, inner sep = .8pt},
    point3/.style = {draw, circle,  fill = red, inner sep = .8pt},
    dot/.style   = {draw, circle,  fill = black, inner sep = .2pt},
  ]
  
   \node (opt) at (18,7.04) [point3, label = above right:$x_{\ast}$] {};
 
   \coordinate (start) at (0,3);
   \node (a1) at (start) [point2, label = above:$\tiny{x_{k-1}}$] {};
   \node (a2) at (4,0) [point, label = below:$x_{k}$] {};
   \node (a3) at (5.5,4.2) [point2, label = above:$x_{k+1}$] {};
   \node (a4) at (9,2.5) [point] {};
   \node (a5) at (11.5,5.7) [point] {};
   \node (a6) at (13.5,5.2) [point] {};
   \node (a7) at (14.3,6.5) [point] {};
   \node (a8) at (15.6,6) [point] {};
   \node (a9) at (16.1,6.8) [point] {};    
   \node (a10) at (16.7,6.5) [point] {};    
   \node (a11) at (17,6.95) [point] {};    
   \node (a12) at (17.35,6.75) [point] {};  
    \node (a13) at (17.5,7.0) [point] {};    
   \node (a14) at (17.7,6.85) [point] {};      
   \node (a15) at (17.8,7.03) [point] {};      
   \node (a16) at (17.92,6.94) [point] {};    

    \node (arrow) at ($(a1)!1.5!(a3)$) {};
   \draw[->,ultra thick] (a1) -- node (a) [above,sloped, pos=0.4] {} (arrow); 
  \draw[dotted] (a1) -- (opt);
  \draw[-] (a1) -- (a2);
  \draw[-] (a2) -- (a3);
  \draw[-] (a3) -- (a4);  
  \draw[-] (a4) -- (a5);  
  \draw[-] (a5) -- (a6);  
  \draw[-] (a6) -- (a7);  
  \draw[-] (a7) -- (a8);  
  \draw[-] (a8) -- (a9);  
  \draw[-] (a9) -- (a10);  
  \draw[-] (a10) -- (a11);  
  \draw[-] (a11) -- (a12); 
  \draw[-] (a12) -- (a13); 
  \draw[-] (a13) -- (a14);   
    \draw[-] (a14) -- (a15);           
        \draw[-] (a15) -- (a16);           
                \draw[-] (a16) -- (opt);      
  \end{tikzpicture}   
  \end{minipage}
\vskip 0.2cm
\caption{\label{zigzag}A geometrical interpretation of the PARTAN-FW iteration.}
\end{center}
\end{figure}
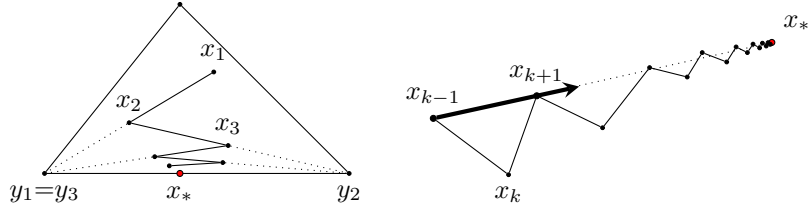

A geometrical interpretation of the PARTAN method can be obtained by looking at the typical behaviour of a standard FW iteration near a solution: the fact that the search direction of the FW method tends to become orthogonal to $\nabla f(\alpha^{(k)})$ close to the optimum can easily lead to a zigzagging trajectory, as seen from Figure \ref{zigzag}(a). A simple way to circumvent this behaviour consists in performing an extra line-search along the line connecting $\alpha^{(k-1)}$ to $\tilde \alpha$ (which corresponds to a basic FW step from $\alpha^{(k)}$). The case depicted in Figure \ref{zigzag}(b) shows how PARTAN is able to avoid traversing the \quotes{sawtooth} in the trajectory, directly moving towards a point closer to the solution. It is apparent how this approach is especially advantageous if the stepsizes can be computed by a closed formula, as is the case, e.g., for quadratic objective functions.

When specialized to the SVM problem (\ref{svm_problem}), the algorithm assumes a simpler form, as the key steps in each iteration can be performed analytically. The necessary formulas, which are obtained via elementary algebraic manipulations, are reported in the Appendix. For the purposes of the discussion here, it suffices to mention that the cost per iteration of the PARTAN method is nearly equivalent to that of the standard FW, as also demonstrated by the numerical results in the next section.
Regarding the stopping criterion, the duality gap can be conveniently computed as
\begin{equation}\label{gap_formula}
\Delta_d^{(k)} = \nabla f(\alpha^{(k)})_{i_*^{(k)}} - 2f(\alpha^{(k)}) \,.
\end{equation}
We summarize the overall procedure in Algorithm \ref{alg:PARTAN}. 

\begin{small}
\begin{algorithm}[!ht]
\begin{algorithmic}[1]
\STATE \textbf{Input:} an initial estimate $\alpha^{(0)}$ and a tolerance $\varepsilon$. \\%
\STATE Compute $\alpha^{(1)}$ via a standard FW step. 
\STATE Search for $i_*^{(1)} \in \argmax_{i} \nabla f(\alpha^{(1)})_i$.
\STATE Initialize the duality gap as in (\ref{gap_formula}).
\STATE Set $k=1$.
\WHILE{$\Delta_d^{(k)} > \epsilon$}
\STATE Compute the optimal FW steplength as in (\ref{fw_stepsize}). 
\STATE Compute the function value after the intermediate FW step as in (\ref{fw_fval}). 
\STATE Compute $W_k$ as in (\ref{recursive_W}). 
\STATE Compute the optimal PARTAN steplength as in (\ref{partan_stepsize}). 
\STATE Perform the PARTAN step (\ref{averaging_step}) as:
\[
\begin{aligned}
\alpha^{(k+1)} = \;&(1 + \mu^{(k)} - \lambda^{(k)} - \mu^{(k)}\lambda^{(k)})\alpha^{(k)} - \\
& \mu^{(k)}\alpha^{(k-1)}  + (\lambda^{(k)} + \lambda^{(k)}\mu^{(k)}) e_{i_*^{(k)}}.
\end{aligned}
\] 
\STATE Update the function value as in (\ref{partan_fval}). 
\STATE Set $k := k+1$. 
\STATE Search for $i_*^{(k)} \in \argmax_{i} \nabla f(\alpha^{(k)})_i$.
\STATE Update the duality gap as in (\ref{gap_formula}). 
\ENDWHILE
\end{algorithmic}
\caption{\label{alg:PARTAN} {The PARTAN-FW algorithm for problem (\ref{svm_problem}).}}
\end{algorithm}
\end{small}

\section{Numerical Results}\label{exp_sec}

In this section, we assess the performance of all the considered variants of FW on the binary classification problem (\ref{svm_problem}), using the benchmark datasets listed in Table \ref{TabDatasets}. The number of examples in the training set and test set are denoted by $m$ and $t$, respectively, while $n$ denotes the number of features.


\begin{table}[h!]
\begin{footnotesize}
\centering
\begin{tabular}{@{}lrrr}
\hline
{Dataset} \T& ${m}$ & $t$  & ${n}$\B\\
\hline
{\textbf{Adult a9a}}  \T   & $32,561$    & $16,281$    & $123$\\
{\textbf{Web w8a}}  & $49,749$  & $14,951$   & $300$\\
{\textbf{IJCNN1}}     & $49,990$     & $91,701$   & $22$\\
{\textbf{USPS-ext}}  & $266,079$    & $75,383$     & $675$\\
{\textbf{KDD99-binary}}   & $395,216$   & $98,805$  & $38$\\
{\textbf{RCV1-binary}}\B    & $677,399$   & $20,242$  & $47,236$\\
\hline
\end{tabular}
\caption{\small \label{TabDatasets} List of the benchmark datasets for problem (\ref{svm_problem}).}
\end{footnotesize}
\end{table}

All the experiments are performed with an RBF kernel. Due to the size of the datasets, the SVM regularization parameter is selected by a simple approach, where a single validation set is built by randomly extracting $70\%$ of the training examples, and the remaining $30\%$ is reserved for testing \footnote{The same values of the hyper-parameters are used for all the methods.}. For the RCV1 dataset, we used the value suggested in \cite{rcv1_paper}. The kernel width is selected according to the heuristic in \cite{coreSVMs05tsang}. The algorithms are coded in C++ and run in Linux on a 3.40 GHz Intel i7 machine with 16 GB of main memory.

In the first experiment, we set $\varepsilon = 10^{-4}$ in the stopping criterion (\ref{duality_gap}) and evaluate the performance of the proposed methods in terms of test accuracy, CPU time (in seconds), number of iterations and model size (number of SVs). Results are reported in Table \ref{exp_1}.

\begin{table}[h!!!]
\centering
\begin{footnotesize}
{\begin{tabular}{@{}lllll}
\hline
\T  & FW & MFW & SWAP & PARTAN \B\\
\hline
\T \textbf{Adult a9a} & & & & \\
   Acc (\%) & $84.21$ & $83.29$ & $83.53$ & $84.00$  \\
 Time  & $1.58\e+02$ & $1.57\e+02$ & $2.26\e+02$ & $1.07\e+02$  \\
 Iter & $2.02\e+04$ & $2.00\e+04$ & $1.76\e+04$ & $1.34\e+04$  \\  
 SVs & $1.39\e+04$ & $1.28\e+04$ & $1.41\e+04$ & $1.18\e+04$   \B \\
\hline
 \T \textbf{Web w8a} & & & & \\
 Acc (\%) & $99.30$ & $99.32$ & $99.28$ & $99.30$ \\
 Time  & $3.78\e+02$ & $3.16\e+02$ & $3.56\e+02$ & $1.07\e+02$  \\
  Iter & $1.65\e+04$ & $1.38\e+04$ & $9.24\e+03$ & $4.62\e+03$  \\  
  SVs & $6.92\e+03$ & $4.48\e+03$ & $4.97\e+03$ & $2.83\e+03$ \B \\
\hline
\T \textbf{IJCNN1} & & & & \\
 Acc (\%) & $98.50$ & $98.22$ & $98.40$ & $98.36$ \\
  Time  & $5.13\e+01$ & $4.57\e+01$ & $5.09\e+01$ & $1.98\e+01$   \\
   Iter & $1.59\e+04$ & $1.41\e+04$ & $1.35e+04$ & $5.48\e+03$  \\  
  SVs & $3.23\e+03$ & $2.73\e+03$ & $3.16\e+03$ & $2.73\e+03$ \B \\
\hline
\T \textbf{USPS-ext} & & & & \\
 Acc (\%) & $99.52$ & $99.52$ & $99.53$ & $99.52$ \\
 Time  & $1.98\e+03$ & $8.44\e+02$ & $9.46\e+02$ & $7.38\e+02$  \\
  Iter & $2.15\e+04$ & $9.17\e+03$ & $8.10\e+03$ & $7.79\e+03$  \\  
  SVs & $3.93\e+03$ & $3.64\e+03$ & $3.67\e+03$ & $3.51\e+03$ \B \\
  \hline
 \T \textbf{KDD99-binary} & & & & \\
 Acc (\%) & $99.94$ & $99.93$ & $99.94$ & $99.93$ \\
 Time  & $7.02\e+02$ & $5.26\e+02$ & $5.51\e+02$ & $1.89\e+02$  \\
  Iter & $1.71\e+04$ & $1.27\e+04$ & $7.82\e+03$ & $4.37\e+03$  \\  
  SVs & $5.25\e+03$ & $3.63\e+03$ & $3.86\e+03$ & $2.82\e+03$ \B \\
\hline
 \T \textbf{RCV1-binary} & & & & \\
 Acc (\%) & $97.55$ & $96.64$ & $97.17$ & $97.50$ \\
 Time  & $1.37\e+04$ & $1.36\e+04$ & $1.71\e+04$ & $1.23\e+04$  \\
  Iter & $3.77\e+04$ & $3.81\e+04$ & $3.65\e+04$ & $3.38\e+04$  \\  
  SVs & $3.75\e+04$ & $3.58\e+04$ & $3.65\e+04$ & $3.37\e+04$ \B \\
 \hline
\end{tabular}}
\caption{\label{exp_1} Comparison of different variants of FW on benchmark SVM datasets.}
\end{footnotesize}
\end{table}

It can be seen that all the algorithms generally exhibit a good performance. In particular, the PARTAN variant in Algorithm \ref{alg:PARTAN} yields the most consistent results, improving on the running times of the plain FW by a factor of $2.52$ on average. Results relative to test accuracy and model sizes are fairly stable, with no particular variant outperforming the others in most cases, though it should be noted that PARTAN is often able to find a smaller SV set. This means that the number of spurious points (i.e. active examples which are not part of the true SV set) selected by the FW iterations is potentially reduced, as especially evident on the \textbf{Web w8a} dataset.

We can also see how the reduction in computational time for PARTAN-FW is roughly proportional to the decrease in the number of iterations, which confirms our intuition that using the PARTAN algorithm on SVMs does not imply a higher iteration complexity than that of the standard FW \footnote{Note that this also holds true for the other variants \cite{SWAP_paper}.}. Seeing how this technique provides a systematic speedup with no evident drawbacks when compared to the standard FW, we recommend it over the latter for large-scale SVM problems.

A potentially relevant observation is that the benefit of using the PARTAN-accelerated iteration is related to a good extent to the sparsity of the solution. The advantage is indeed more apparent on problems where the size of the SV set is a small fraction of the total number of examples, with the \textbf{KDD99-binary} dataset being a prominent example.

On the other hand, the FW variants show no advantage over the standard algorithm on the \textbf{RCV1-binary} problem. This is arguably because the number of SVs is basically the same as the total number of iterations. Since the FW algorithm spends all of its iterations adding new vertices (i.e. examples corresponding to nonzero components of $\alpha^{(k)}$) to the model, the usual slowdown behaviour of FW, where the algorithm 
cycles between the same vertices readjusting their weights, is not observed. As such, there is little benefit in adding modified FW directions. The same phenomenon is observed, on a smaller scale, on the \textbf{Adult a9a} dataset.

This is consistent with the fact that FW methods 
are best used to solve sparse problems, as also suggested by their theoretical properties. Conversely, their usefulness is more limited when the solution is dense, as the incremental nature of the algorithm provides no particular advantage in this case.

As far as the difference in performance between the variants is concerned, we remark that the theoretical results in Section \ref{intro-sec} are of asymptotic nature, and as such it is difficult to assess their practical impact with a fixed value of $\varepsilon$, which might be too large to observe a faster convergence compared to the standard FW. We investigate this issue in the next paragraph.

\subsection{Considerations on the Iteration Complexity}

We now attempt to better assess the practical difference between the standard FW and its variants, in order to understand how and when the latter can give a subtantial advantage.
In particular, we want to estabilish whether the improved convergence predicted by the theory can be observed experimentally when using a duality gap-based stopping criterion.
To this end, we apply all the considered variants of FW to the datasets \textbf{Adult a9a}, \textbf{Web w8a} and \textbf{IJCNN1}, using increasingly strict tolerances $\varepsilon \in \{10^{-3},\ldots,10^{-6}\}$, and monitoring the number of iterations needed to trigger the stopping condition. We do not attempt to solve the larger scale problems here, as the smallest value of $\varepsilon$ would lead to prohibitive running times, and we remark that this experiment aims exclusively at providing an insight on the convergence speed of the algorithms. Results are shown in Table \ref{exp_2}.

\begin{table}[h!!!]
\begin{footnotesize}
\centering
{\begin{tabular}{@{}llllll}
\hline
\T & $\varepsilon$ & $1\e-03$ & $1\e-04$ & $1\e-05$ & $1\e-06$ \B\\
\hline
\T \textbf{Adult a9a}\!\!\! & & & & \B \\
\T {FW} & Time\!\!  & $2.24\e+01$\! & $1.58\e+02$\!\! & $1.46\e+03$\!\! & $1.42\e+04$\!\!\!\!\!  \\
 & Iter & $2.82\e+03$\! & $2.02\e+04$\!\! & $1.84\e+05$\!\! & $1.79\e+06$\!\!\!\!\!  \\  
 \T {MFW} & Time\!\! & $2.20\e+01$\!\! & $1.57\e+02$\!\! & $5.48e+02$\!\! & $1.21\e+03$\!\!\!\!\!  \\
  & Iter & $2.77\e+03$\!\! & $2.00\e+04$\!\! & $6.80\e+04$\!\! & $1.50\e+05$\!\!\!\!\! \\  
 \T {SWAP} & Time\!\!  & $3.07\e+01$\!\! & $2.26\e+02$\!\! & $6.46\e+02$\!\! & $1.30\e+03$\!\!\!\!\!  \\
  & Iter & $2.61\e+03$\!\! & $1.76\e+04$\!\! & $5.18\e+04$\!\! & $1.09\e+05$\!\!\!\!\!  \\  
 \T {PARTAN}\!\!  & Time\!\!  & $1.73\e+01$\!\! & $1.07\e+02$\!\! & $5.09\e+02$\! & $5.43\e+03$\!\!\!\!\!  \\
  & Iter & $2.15\e+03$\!\! & $1.64\e+04$\!\! & $6.21\e+04$\!\! & $6.63\e+05$\!\!\!\!\!  \B \\  
\hline
\T \textbf{Web w8a}\!\!\! & & & & \B \\
\T {FW} & Time\!\!  & $4.47\e+01$\!\!\! & $3.78\e+02$\!\!\! & $3.73\e+03$\!\!\! & $4.02\e+04$\!\!\!\!\!  \\
 & Iter & $1.93\e+03$\!\! & $1.65\e+04$\!\! & $1.63\e+05$\!\! & $1.75\e+06$\!\!\!\!\!  \\  
 \T {MFW} & Time\!\! & $4.27\e+01$\!\! & $3.16\e+02$\!\! & $1.52\e+03$\!\! & $3.78\e+03$\!\!\!\!\!  \\
  & Iter & $1.86\e+03$\!\! & $1.38\e+04$\!\! & $6.38\e+04$\!\! & $1.65\e+05$\!\!\!\!\! \\  

 \T {SWAP} & Time\!\!  & $6.42\e+01$\!\! & $3.56\e+02$\!\! & $1.31\e+03$\!\! & $8.63\e+03$\!\!\!\!\!  \\
  & Iter & $1.69\e+03$\!\! & $9.24\e+03$\!\! & $4.29\e+04$\!\! & $3.46\e+05$\!\!\!\!\!  \\  
 \T {PARTAN}\!\!  & Time\!\!  & $1.83\e+01$\!\! & $1.07\e+02$\!\! & $5.97\e+02$\!\! & $3.32\e+03$\!\!\!\!\!  \\
  & Iter & $7.77\e+02$\!\! & $4.62\e+03$\!\! & $2.57\e+04$\!\! & $1.44\e+05$\!\!\!\!\!  \B \\  
\hline
\T \textbf{IJCNN1} & & & & & \B \\
\T {FW} & Time\!\! & $5.77\e+00$\!\! & $5.13\e+01$\!\! & $5.49\e+02$\!\! & $6.48\e+03$\!\!\!\!\!  \\
 & Iter & $1.72\e+03$\!\! & $1.59\e+04$\!\! & $1.68\e+05$\!\! & $1.97\e+06$\!\!\!\!\!  \\  
\T {MFW} & Time\!\! & $5.16\e+00$\!\! & $4.57\e+01$\!\! & $2.38\e+02$\!\! & $6.81\e+02$\!\!\!\!\!  \\
  & Iter & $1.51\e+03$\!\! & $1.41\e+04$\!\! & $7.12\e+04$\!\! & $2.05\e+05$\!\!\!\!\!  \\  
\T {SWAP} & Time\!\! & $6.94\e+00$\!\! & $5.09\e+01$\!\! & $2.56\e+02$\!\! & $6.56\e+02$\!\!\!\!\!  \\
 &  Iter & $1.21\e+03$\!\! & $1.35e+04$\!\! & $6.85\e+04$\!\! & $1.77\e+05$\!\!\!\!\!  \\  
\T {PARTAN}\!\! & Time\!\!  & $3.09\e+00$\!\! & $1.98\e+01$\!\! & $1.60\e+02$\!\! & $1.83\e+03$\!\!\!\!\!   \\
  & Iter & $7.83\e+02$\!\! & $5.48\e+03$\!\! & $4.34\e+04$\!\! & $4.99\e+05$\!\!\!\!\!\B \\  
\hline
\end{tabular}}
\caption{\label{exp_2} Iteration complexity of different variants of FW.}
\end{footnotesize}
\end{table}

From the results, it is clear how the standard FW behaves according to the $\cO(1/\varepsilon)$ iteration bound, with the number of iterations increasing 10-fold every time the tolerance parameter decreases by one order of magnitude. This corresponds to the duality gap decreasing as $\cO(1/k)$, as predicted by Proposition \ref{dualconv}. 
This result suggests that, though (\ref{dual_bound}) is an upper bound of the duality gap (and thus in turn of the primal gap), it gives in practice a good indication of the number of iterations that we can expect from the standard FW algorithm, therefore implying that the computational effort can be predicted and controlled by appropriately tuning the tolerance parameter.
The modified variants, in contrast, enjoy a faster convergence rate, ending up gaining a computational advantage of one order or magnitude or more with respect to the plain FW when the strictest tolerance value is used. 
It is interesting to note how the three variants analyzed here do not always provide the same improvement. As the results presented in this work are only preliminary, it is difficult to establish whether this is simply due to the methods having different convergence factors (e.g. because they enjoy convergence rates with the same asymptotic behaviour but differing by a constant) or is intrinsically related to the nature of the algorithms (for example, PARTAN starts out with a substantial advantage over the other variants at $\varepsilon = 10^{-4}$, but is outperformed by MFW when seeking for a more accurate solution).

\begin{figure}[h!!!]
\begin{center}
\subfigure[]{
{\includegraphics[scale = 0.095]{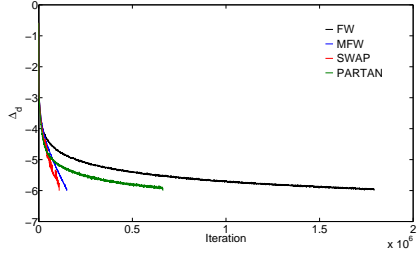}}}
\subfigure[]{
   {\includegraphics[scale = 0.095]{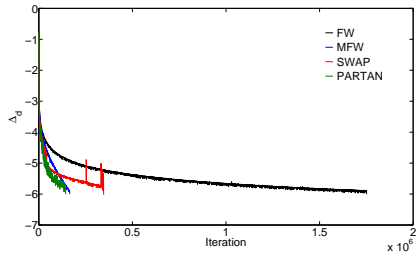}}}
\subfigure[]{
   {\includegraphics[scale = 0.095]{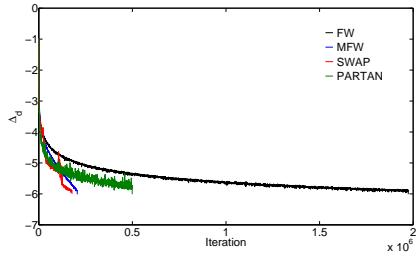}}}
\caption{\label{dual_path} Duality gap behaviour of FW algorithms for $\varepsilon = 10^{-6}$ on the datasets \textbf{Adult a9a} (a), \textbf{Web w8a} (b), and \textbf{IJCNN1} (c).} 
\end{center}
\end{figure}

We can attempt to shed some more light on this issue by plotting in Figure \ref{dual_path} the duality gap (in logarithmic scale), obtained with $\varepsilon = 10^{-6}$, against the iteration number for all the variants of the algorithm. From the graphs, it can be seen how the MFW algorithm seems to exhibit the best primal-dual convergence rate, with oscillations in the duality gap being very small \footnote{It is important to observe that, as opposed to $f(\alpha^{(k)})$, $\Delta_d^{(k)}$ is not a monotonically decreasing quantity.}. The SWAP algorithm performs even better on two of the datasets (though exhibiting larger oscillations), but substantially worse on \textbf{Web w8a}. The PARTAN variant seems instead to show a behaviour similar to that of the standard FW, but with a better convergence factor, an observation which is consistent with the results obtained in Tables \ref{exp_1} and \ref{exp_2}. It might also be worth noticing that $\Delta_d^{(k)}$ is only an upper bound of the optimality measure $f(\alpha^{(k)})-f(\alpha^*)$, thus occasional oscillating values of the duality gap do {not} imply that the solution is getting less accurate.

Overall, though we are well aware that a 
more representative batch of problems would be needed to draw more solid conclusions, the results in Table \ref{exp_2} show that the considered FW variants indeed display a faster convergence rate in practice using a primal-dual stopping criterion. It is not obvious, however, whether the results on the duality gap given by Proposition \ref{dualconv} can be improved under suitable hypotheses.
They also show how the traditional FW is not a suitable method if one wants to use stricter values of $\varepsilon$, for example because the application at hand requires a higher optimization accuracy. This confirms on a Machine Learning problem the well-known intuition that the standard FW step stagnates when close to a solution, unless extra search directions which do not get orthogonal to the gradient are added \cite{IJPRAI11}.

It should be noted, indeed, that a good choice of $\varepsilon$ is application-dependent. In SVMs for classification, for instance, it is well known that 
the test accuracy is often relatively insensitive to $\varepsilon$ after a certain threshold. On the other hand, different applications, such as function estimation, could be more sensitive to the accuracy (in an optimization sense) of the obtained model.
Therefore, while they may not appear very relevant in the context of classification SVMs, the improved properties of some FW modifications may be of importance for other related tasks. In this case, we would recommend the use of a FW variant rather than the standard algorithm. 

\subsection{Results with Randomized Iterations}

As the total number of iterations required by a FW algorithm can be large, devising a convenient way to solve the subproblem (\ref{linear_subpb}) is recommended in order to make the algorithm more viable on large-scale datasets. A typical situation arises when (\ref{linear_subpb}) has an analytical solution or it is easy to solve due to the problem structure  \cite{SWAP_paper, rinaldi13}. This is the case, for example, for all the problems introduced in Section \ref{applications_sec}. Still, the resulting complexity usually depends on the problem size (for example, in (\ref{svm_vertex}) it is proportional to $m$), and can thus be impractical when handling large-scale data.

A simple and yet effective way to avoid the dependence on $m$ is to look for the solution of (\ref{linear_subpb}) by exploring only a fixed number of extreme points on the boundary of $\Sigma$ \cite{coreSVMs05tsang,Smola01Learning,NIPS14}. 
In the case of (\ref{svm_problem}), for example, this means extracting a sample $\mathcal{S} \subseteq \{1,\ldots,m\}$ and solving
\begin{equation}\label{sampling_iter}
i^{(k)}_\mathcal{S} \in \argmin_{i \in \mathcal{S}} \nabla f(\alpha^{(k)})_i \,.
\end{equation}
The cost of an iteration becomes in this case $\mathcal{O}(|\mathcal{S}||\mathcal{I}^{(k)}|)$, rather than $\cO(m|\cI^{(k)}|)$ as in (\ref{svm_vertex})\footnote{It should also be noted that a clever implementation allows to eliminate the $|\cI^{(k)}|$ factor from (\ref{svm_vertex}) in the SVM case \cite{NIPS14}.}.

The stopping criterion, however, is not applicable without computing the entire gradient $\nabla f(\alpha^{(k)})$, which is not done in the randomized case. As a possible alternative, we can use the approximate quantity
\[
\Delta_\mathcal{S}(\alpha^{(k)}) := 2 f(\alpha^{(k)}) - \nabla f(\alpha^{(k)})_{i^{(k)}_\mathcal{S}}.
\]
Since $\Delta_\mathcal{S}(\alpha^{(k)}) \leq \Delta_d^{(k)}$, this simplification entails a tradeoff between the reduction in computational cost and risk of a premature stopping. Although this can be acceptable in contexts such as SVM classification, where solving the optimization problem with a high accuracy is usually not needed, it is important to make sure that the impact of this approximation is kept to an acceptable level. This issue has been discussed in detail in \cite{NIPS14}.
Here, to mitigate the effect of a possible early stopping, we implement a simple safeguard strategy where the sampling (\ref{sampling_iter}) is repeated twice in case $\Delta_\mathcal{S}(\alpha^{(k)}) \leq \varepsilon$.

In Table \ref{exp_random}, we report the results obtained with a randomization technique, taking $|\mathcal{S}| = 194$ \footnote{This value corresponds to a probability of at least $0.98$ that $i^{(k)}_*$ lies in the $2\%$ smallest components of $\nabla f(\alpha^{(k)})$. See Theorem 6.33 in \cite{Smola01Learning}.}, averaged over $10$ runs. Note that we do not attempt to run the randomization technique on problems for which this strategy is not beneficial. Taking \textbf{RCV1-binary} as an example, it is already clear, from the structure of the dataset and the results in Table \ref{exp_1}, that using a random sampling would not provide any advantage: the SV set size being of order $10^{4}$, an iteration would have a complexity in the order of millions of floating point operations, which is actually much larger than the size of the whole dataset. In general, we do not recommend using a random sampling for problems with dense SV sets, as in order to obtain a computational gain the number of samples would have to be too small, possibly leading to an inaccurate solution.

\begin{table}[h!!!]
\begin{footnotesize}
\centering
{\begin{tabular}{@{}lllll}
\hline
\T &  FW & MFW & SWAP & PARTAN \B\\
\hline
\T \textbf{Adult a9a} & & & & \\
  Acc (\%) & $83.94$ & $83.86$ & $83.52$ & $84.08$  \\
 Time  & $1.69\e+02$ & $1.63\e+02$ & $1.64\e+02$ & $1.08\e+02$  \\
  Iter & $1.89\e+04$ & $1.98\e+04$ & $1.65\e+04$ & $1.24\e+04$  \\  
  SVs & $1.35\e+04$ & $1.24\e+04$ & $1.35\e+04$ & $1.11\e+04$   \B \\
\hline
 \T \textbf{Web w8a} & & & & \\
 Acc (\%) & $99.17$ & $99.21$ & $99.16$ & $99.10$ \\
 Time  & $8.39\e+01$ & $6.69\e+01$ & $7.12\e+01$ & $4.89\e+01$  \\
  Iter & $6.78\e+03$ & $9.10\e+03$ & $4.97\e+03$ & $3.25\e+03$  \\  
  SVs & $3.66\e+03$ & $3.01\e+03$ & $3.19\e+03$ & $2.31\e+03$ \B \\
\hline
\T \textbf{IJCNN1} & & & & \\
Acc (\%) & $98.57$ & $97.98$ & $98.34$ & $$98.37 \\
 Time  & $3.45\e+01$ & $2.10\e+01$ & $2.21\e+01$ & $1.95\e+01$   \\
   Iter & $1.12\e+04$ & $1.14\e+04$ & $7.10\e+03$ & $5.76\e+03$  \\  
  SVs & $4.12\e+03$ & $2.45\e+03$ & $3.43\e+03$ & $3.34\e+03$ \B \\
\hline
\T \textbf{USPS-ext} & & & & \\
Acc (\%) & $99.53$ & $99.57$ & $99.55$ & $99.53$ \\
 Time  & $2.19\e+02$ & $2.60\e+02$ & $2.25\e+02$ & $2.07\e+02$  \\
  Iter & $3.61\e+03$ & $6.49\e+03$ & $3.59\e+03$ & $3.28\e+03$  \\  
  SVs & $2.96\e+03$ & $2.48\e+03$ & $2.94\e+03$ & $2.86\e+03$ \B \\
 \hline
 \T \textbf{KDD99-binary} & & & & \\
Acc (\%) & $99.73$ & $99.93$ & $99.78$ & $99.82$ \\
 Time (s) & $2.84\e+01$ & $9.46\e+01$ & $2.03\e+01$ & $1.20\e+01$  \\
  Iter & $1.88\e+03$ & $1.29\e+04$ & $1.57\e+03$ & $1.15\e+03$  \\  
  SVs & $1.71\e+03$ & $2.35\e+03$ & $1.45\e+03$ & $1.11\e+03$ \B \\
 \hline
\end{tabular}}
\caption{\label{exp_random} Comparison of different variants of FW on benchmark SVM datasets (randomized iteration).}
\end{footnotesize}
\end{table}

First of all, note that the effect of sampling is substantially problem-dependent. The best computational gains are obtained on the problems \textbf{Web w8a}, \textbf{USPS-ext} and \textbf{KDD99-binary}, with the latter two being the largest and most sparse datasets. It should be noted that the reduction in CPU time is attributable both to the reduced iteration complexity and to the smaller iteration count, the latter being due to the approximate stopping criterion employed. This is not observed on all the datasets, however. For example, the total number of iterations on \textbf{Adult a9a} and \textbf{IJCNN1} is comparable to that of the deterministic case. 
The MFW algorithm also appears to be overall less sensitive to this particular issue.

It is interesting to note that this phenomenon does not necessarily lead to a loss in test accuracy. This is possibly due to the nature of the SVM classification models, which do not require a very accurate solution of the optimization problem to build a decision function with a good predictive capability.


\section{Conclusions}\label{conclusions_sec}
The results presented in this paper show that the family of FW algorithms obtains promising results on several benchmark SVM classification tasks, offering a solid and fast alternative to the classical solvers used in this field. 

While the experimental results presented here are preliminary, they provide the first example of a successful application of the PARTAN-FW iteration to Machine Learning problems, showing that this variant is able to accelerate the basic FW iteration in a systematic way. On the other hand, the advantage of other 
modified FW algorithms is especially apparent when employing stricter tolerance parameters, arguably due to the stronger theoretical properties of the enhanced iterations.

Finally, we have shown how, on larger scale problems, a randomization technique can be employed to reduce the computational effort with satisfactory results, with some caveats related to the tradeoff between complexity and optimization accuracy which is inherent to this kind of strategy.

Experiments on different machine learning applications (such as Lasso and matrix recovery problems) are currently being investigated, and will be the subject of another paper.

\subsection*{Acknowledgments}
\begin{footnotesize}
The research leading to these results has received funding from the European Research Council under the European Union's Seventh Framework Programme (FP7/2007-2013) / ERC AdG A-DATADRIVE-B (290923). This paper reflects only the authors' views and the Union is not liable for any use that may be made of the contained information. Research Council KUL: GOA/10/09 MaNet, CoE PFV/10/002 (OPTEC), BIL12/11T; Flemish Government: FWO: projects: G.0377.12 (Structured systems), G.088114N (Tensor based data similarity); PhD/Postdoc grants; iMinds Medical Information Technologies SBO 2014; IWT:  POM II SBO 100031; Belgian Federal Science Policy Office: IUAP P7/19 (DYSCO, Dynamical systems, control and optimization, 2012-2017).
The second author received funding from CONICYT Chile through FONDECYT Project 11130122.
\end{footnotesize}

\section*{Appendix: Implementation Details}
We report here, for the sake of completeness, the analytical formulas used in the implementation of Algorithm \ref{alg:PARTAN} for the SVM problem (\ref{svm_problem}). To simplify the equations, we use the shorthand notations $f^{(k)} = f(\alpha^{(k)})$ and $ \nabla f^{(k)} = \nabla f(\alpha^{(k)})$.

After some elementary algebraic manipulations, we obtain that the optimal steplength value for 
step (\ref{intermediate_step}) is given by
\begin{equation}\label{fw_stepsize}
\lambda^{(k)} = \frac{2f^{(k)} - \nabla f^{(k)}_{i_*^{(k)}}}{2f^{(k)} - 2 \nabla f^{(k)}_{i_*^{(k)}} - K_{i_*^{(k)},i_*^{(k)}}}. 
\end{equation}
After this step, the objective value becomes 
\begin{equation}\label{fw_fval}
\begin{aligned}
\tilde f = \;& (1 - \lambda^{(k)})^2 f^{(k)} + \lambda^{(k)}(1-\lambda^{(k)}) \nabla f^{(k)}_{i_*^{(k)}} \\
& - \tfrac{1}{2} (\lambda^{(k)})^2 K_{i_*^{(k)},i_*^{(k)}}.
\end{aligned}
\end{equation}
The steplength for the PARTAN step (\ref{averaging_step}) is then given by
\begin{equation}\label{partan_stepsize}
\mu^{(k)} = \frac{\lambda^{(k)} \nabla f^{(k-1)}_{i_*^{(k)}} - (1-\lambda^{(k)}) W^{(k)} - 2\tilde f}{2( \tilde f+ (1-\lambda^{(k)}) W^{(k)} - \lambda^{(k)} \nabla f^{(k-1)}_{i_*^{(k)}} + f^{(k-1)}) },
\end{equation} 
where
\begin{equation}\label{recursive_W}
\begin{aligned}
W^{(k)} =& -2 (1+\mu^{(k-1)})(1 - \lambda^{(k-1)}) f^{(k-1)}\\
&-  (1+\mu^{(k-1)})\lambda^{(k-1)} \nabla f^{(k-1)}_{i_*^{(k-1)}} -  \mu^{(k-1)} W^{(k-1)}
\end{aligned} 
\end{equation}
is a quantity that can be computed recursively starting from $W^{(1)} = (\alpha^{(0)})^T K \alpha^{(1)}$.
Finally, the updated objective value after the PARTAN iteration is
\begin{equation}\label{partan_fval} 
\begin{aligned}
f^{(k+1)} =\; &
(1+\mu^{(k)})^2 \tilde f 
+ \mu^{(k)} (1+\mu^{(k)}_{k}) (1 - \lambda^{(k)}) W_k \\
&- \mu^{(k)} (1+\mu^{(k)})  \lambda^{(k)} \nabla f^{(k-1)}_{i_*^{(k)}} +  (\mu^{(k)})^2 f^{(k-1)}.
\end{aligned}
\end{equation}

\bibliographystyle{IEEEtran}
\bibliography{PARTAN_bibliography}

\end{document}